\documentclass[10pt,twocolumn,letterpaper]{article}

\usepackage{cvpr}
\usepackage{times}
\usepackage{epsfig}
\usepackage{graphicx}
\usepackage{amsmath}
\usepackage{amssymb}
\usepackage{lipsum}
\usepackage{epigraph}
\usepackage{subcaption}
\usepackage{multirow}
\usepackage{graphicx}
\usepackage[dvipsnames]{xcolor}

\usepackage[pagebackref=true,breaklinks=true,letterpaper=true,colorlinks,bookmarks=false]{hyperref}
\usepackage{authblk}

\usepackage[
  separate-uncertainty = true,
  multi-part-units = repeat
]{siunitx}

 \cvprfinalcopy 

\def\cvprPaperID{17} 

\ifcvprfinal\fi
\begin{document}

\title{Lifting Monocular Events to 3D Human Poses}

\author{Gianluca Scarpellini$^{1,2}$, Pietro Morerio$^{1}$, Alessio Del Bue$^{1,3}$  \\
$^{1}$Pattern Analysis \& Computer Vision - Istituto Italiano di Tecnologia, Italy\\
$^{2}$University of Genova, Genoa, Italy\\
$^{3}$Visual Geometry and Modelling, Istituto Italiano di Tecnologia, Italy \\
{\tt\small \{gianluca.scarpellini,pietro.morerio,alessio.delbue\}@iit.it}
}
\maketitle

\begin{abstract}

This paper presents a novel 3D human pose estimation approach using a single stream of asynchronous events as input. Most of the state-of-the-art approaches solve this task with RGB cameras, however struggling when subjects are moving fast. On the other hand, event-based 3D pose estimation benefits from the advantages of event-cameras, especially their efficiency and robustness to appearance changes. Yet, finding human poses in asynchronous events is in general more challenging than standard RGB pose estimation, since little or no events are triggered in static scenes. Here we propose the first learning-based method for 3D human pose from a single stream of events. Our method consists of two steps. First, we process the event-camera stream to predict three orthogonal heatmaps per joint; each heatmap is the projection of of the joint onto one orthogonal plane. Next, we fuse the sets of heatmaps to estimate 3D localisation of the body joints. As a further contribution, we make available a new, challenging dataset for event-based human pose estimation by simulating events from the RGB Human3.6m dataset. 
Experiments demonstrate that our method achieves solid accuracy, narrowing the performance gap between standard RGB and event-based vision. The code is freely available at \small \url{https://iit-pavis.github.io/lifting_events_to_3d_hpe}. 

\end{abstract}

\section{Introduction}
\begin{figure}[t] \centering
    \includegraphics[width=.93\linewidth]{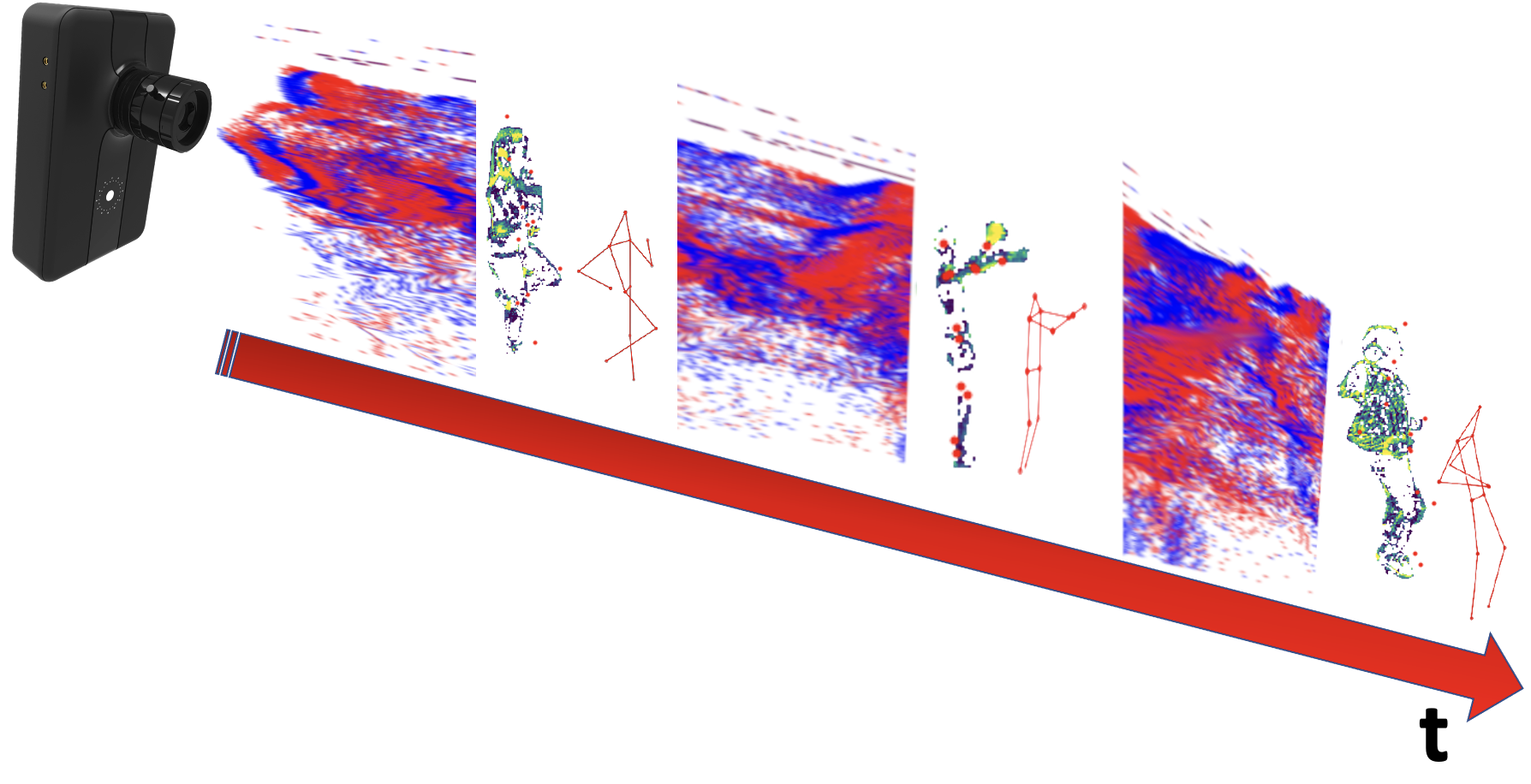}
  \caption{\label{fig:proposal} Our method computes the 3D pose of a subject from event-camera streams. We first aggregate events into meaningful representations that are then used to estimate the final 3D pose of the subject.} 
\end{figure}
Natural selection has empowered us with an efficient perception system, enabling
our brain to process visual information and respond to threats promptly. Biological evidence
suggests that humans and other animals process visual cues differently from
traditional cameras \cite{Mahowald_1991,Zeki_2015}. Instead of handling frames at fixed time intervals, mammals
collect visual cues asynchronously and elaborate information on demand. This observation pushed the research community and engineers to develop new sensors, event-cameras, with a neuromorphic inspiration that provide crucial advantages in time-critical tasks and applications \cite{Lichtsteiner_2008, Brandli:200837}.

Indeed, one of the most important activities we are daily involved in is interacting with other human beings. For this reason, we developed the ability to forecast human motion and adapt our behavior accordingly \cite{scheflen1972body, knapp2013nonverbal, 8883081}. 
%
However, in order to encode asynchronous quick reactions to human activities, a basic but fundamental task to solve is the estimation of human pose from event-based streams. 
Human pose estimation is already widely adopted in action recognition \cite{li17_skelet_cnn, li18_skelet_based_action_recog_by}, human tracking \cite{li17_skelet}, sport assistance \cite{wang_2019}, and virtual reality \cite{vicon}. Most of the adopted solutions involve using multiple cameras and require the subjects to wear special markers suites \cite{vicon}. Despite their efficiency and broad adoption, these techniques rely on delicate synchronization and are difficult to deploy in real environments. For these reasons, monocular human pose estimation represents a fascinating research challenge with growing interests in the industry \cite{comport2003real,shingade2014animation,rosenhahn2008markerless}. There are two different families of solutions to solve 3D human pose estimation: skeleton-based and model-based solutions. The former regresses skeletal 3D joints from a planar image \cite{Mehta2017VNect,Nibali20183DHeatmaps,pavlakos17_coars_fine_volum_predic_singl}, while the latter fits a tri-dimensional parametric model of the human body to the subjects in the scene \cite{hassan2019resolving,bogo2016keep}. Recently, Xu \etal adapted a model-based approach to event-cameras \cite{xu20_event}. Although they underline an interesting solution, their approach requires RGB images to guide the tracking and cannot be used for real-time applications, as it relies on a time-consuming optimization phase.

We propose instead an event-only approach to predict skeletal poses from a single stream of events (Figure \ref{fig:proposal}). Our pipeline consists of two steps. 
First, a Convolutional Neural Network predicts the projection of each joint of the skeleton onto three orthogonal planes. Instead of predicting the positions directly, we constrain our approach onto estimating intermediate heatmaps of probabilities for each joint. Second, we triangulate the sets of 2D positions of each joint to predict the 3D joint pose. 
Similar workss adopt raw events to solve pose estimation tasks \cite{valeiras16_neurom_event_based_pose_estim,xu20_event}. On the other hand, we aggregate events into tensor-like representations. Although event-representations have been widely investigated and validated \cite{Sironi2018HATS:Classification, Lagorce2017HOTS:Recognition, 10.1007/978-3-030-11024-6_54, gehrig2019end}, no previous work has explored these approaches for monocular human pose estimation. Moreover, differently from standard computer vision, where transfer learning across different tasks has been widely investigated \cite{taskonomy2018}, it is still unclear whether pre-training on related tasks can improve event-based human pose estimation. To fill these gaps, we compare different pre-training tasks and different event-representations.

Experiments on natural and synthetic events validate our approach. For validating performance on real event-camera recordings, we adopt the recent DHP19 dataset \cite{Calabrese_2019_CVPR_Workshops}. DHP19 provides recordings of 33 activities from four different points of view. Despite the excellent contribution to event-based vision, DHP19 provides few self-occlusions or hard situations, as most of the activities are conducted on the spot. To fill these gaps, we propose a new, challenging event-based dataset for Human Pose Estimation by simulating events from the standard Human3.6M dataset \cite{ionescu14_human}. The event-camera community proposed numerous simulation tools to tackle the absence of data \cite{Rebecq2018ESIM:Simulator}, and these solutions have been successfully adopted in recent work \cite{rebecq20_high_speed_high_dynam_range}. Human3.6m provides challenging scenarios, such as people walking and moving extensively in the scene, that are intrinsically harder. In fact, we test our proposal on both DHP19 and Event-Human3.6 and provide different ablations and experiments to support our claims. To summarize, our proposal consists of three main contributions:
\begin{itemize}
\setlength\itemsep{0pt}
    \item A pipeline to predict human poses from a single stream of events;
    \item A new, synthetic dataset for benchmarking event-based Human Pose Estimation;
    \item Extensive experiments to validate transfer learning and pre-training approaches for event-based human pose estimation.
    
\end{itemize}

\section{Related work}
In this Section, we discuss skeleton-based approaches for solving monocular Human Pose Estimation and underline their critical points. To solve the limitations, previous works have focused on high-speed cameras; these approaches suffer, however, from high computational and storage limitations. Event-cameras can be a solution to these problem. Indeed, recent works have adopted event-by-event approaches to track objects and subjects in real-time. On the other hand, our approach aggregates events into tensor-like representation, which can be fed to standard Deep Learning models. Moreover, we recognize a gap of challenging datasets for event-based human pose estimation and discuss events simulation and its benefits.

\begin{figure*}[ht] \centering
    \includegraphics[width=.68\textwidth]{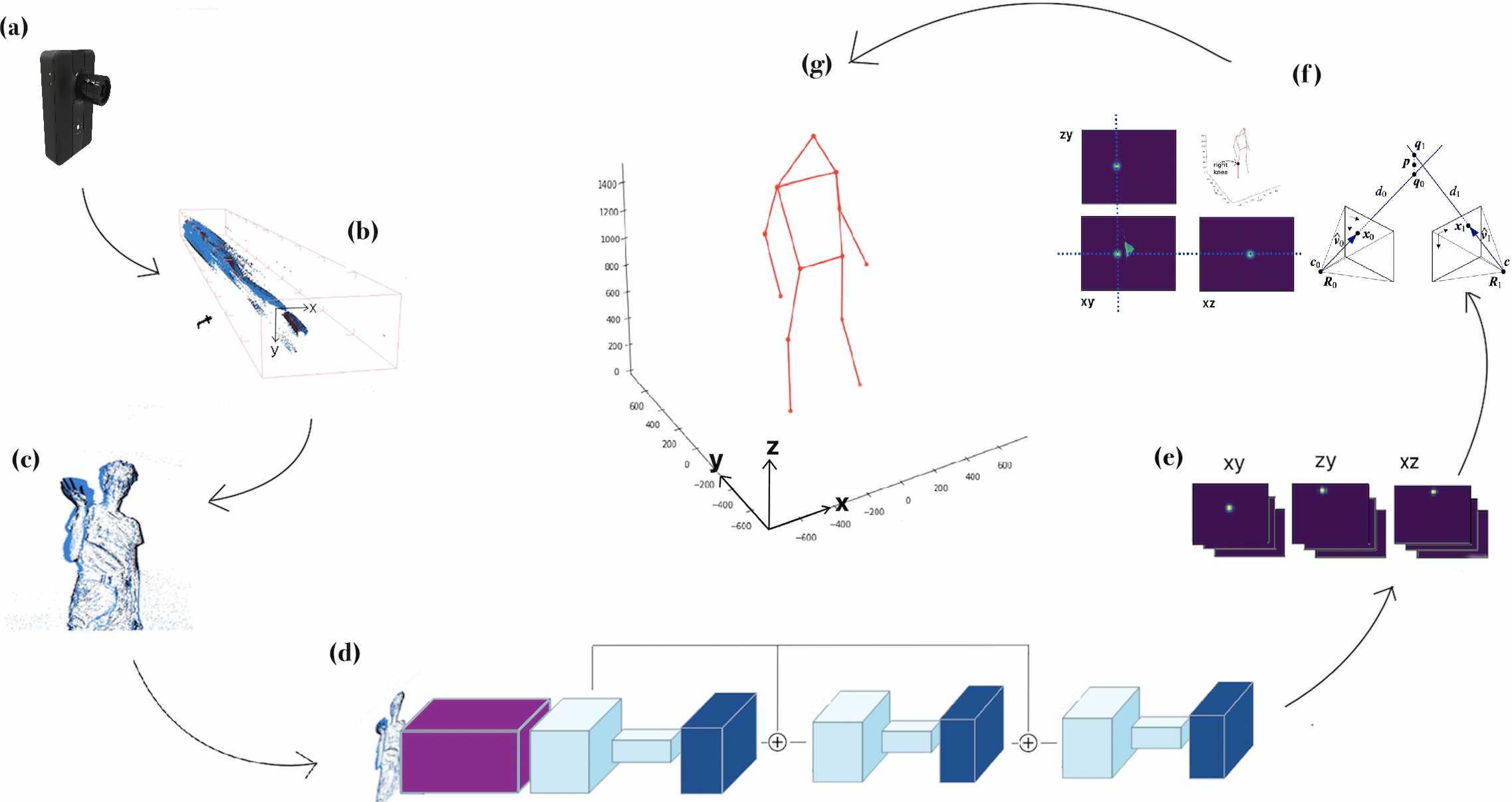}
  \caption[Overview of our approach]{ \label{fig:abstract} \textbf{(a)} A moving subject is recorded with an event-camera. \textbf{(b)} The recording is an asynchronous train of events; each event is characterized by an image plane coordinate (x, y), a timestamp (t), and a positive or negative polarity (respectively, blue and black in the figure). \textbf{(c)} batches of events are accumulated to build frames. Our model processes frames of events \textbf{(d)} in multiple stages. The model output are three set of independent planes; each subject's joint (i.e., head, left and right wrists, and so on) is onto three independent planes\textbf{(e)}. Next, it triangulates the planar predictions \textbf{(f)} and estimates the position of each joint. The output \textbf{(g)} is the subject's skeleton in tridimensional coordinate.}
\end{figure*}

\textbf{Monocular Human Pose Estimation.} Industry and academia are looking at human pose estimation with increasing interest \cite{comport2003real, shingade2014animation, rosenhahn2008markerless}. Commercial solutions usually require special markers suite to track subjects from multiple point of views \cite{vicon}. Despite their satisfactory performances, these approaches are extremely costly and require careful setup choices to perform well at high speeds \cite{merriaux2017study}. 
For these reasons, monocular approaches have been widely researched \cite{xu18_monop}. Along with background, light conditions, texture, and image imperfection, monocular solutions must also handle the intrinsic ambiguity of monocular vision and therefore pose unanswered challenges to the research community. Model-based solutions are an established line of work on this problem. These approaches estimate the full 3D body and shape of the subjects by fitting a model of human body \cite{loper15_smpl}. Although recent model-based works achieve impressive performances \cite{hassan2019resolving,bogo2016keep,huang2017towards}, here we will focus on skeleton-based solutions. Skeleton-based approaches aim to regress 3D joints of the skeleton directly from images. As machine learning models deal with probability better than with scalar values, recent solutions predict dense probability maps (denominated heatmaps) of the location of the skeletal joints onto the image plane. In particular, Newell \etal made a break-trough in the field by proposing Stacked-hourglass model \cite{Newell2016StackedEstimation}. The authors stack multiple Convolutional Neural Networks to extract expressive heatmaps and apply a differentiable sort-argmax operator to retrieve the 2D pixel location of each joint. 
Although we can adapt stacked-hourglass models to predict 3D (volumetric) heatmaps \cite{pavlakos17_coars_fine_volum_predic_singl}, this path is widely open for improvements, especially since Volumetric Heatmaps are computational and memory demanding \cite{Luvizon_2D3D_Pose_Estimation_CVPR_2018_paper}. 
Indeed, Mehta \etal factorize volumentric heatmaps into three 2D heatmaps to lower computational costs \cite{Mehta2017VNect}. The authors train a deep learning model (VNect) to predict x, y, and z axes as dense 2D heatmaps and combine the predictions through triangulation. The computational and resources savings of VNect come with a price in terms of accuracy, as this method reaches higher Mean Per-Joint Precision Error (MPJPE) on common benchmarks. 
Nibali \etal develop this approach further and propose a model (Margipose) to predict $xy$, $zy$, and $xz$ heatmaps and regress the final 3D pose \cite{Nibali20183DHeatmaps}. Others advancements in 3D human pose estimation include GANs \cite{Chen_2019} and temporal convolutions \cite{Cheng_2020}.

\textbf{Event-based approaches.} Real-time applications require a careful design to meet strong computational, speed, and energy requirements. This premise is especially true when fast-moving human subjects are involved, such as in sport assistance and virtual reality. Monocular solutions involving RGB-D sensors \cite{yuan19_tempor_upsam_depth_maps_using_hybrid_camer} and high-speed cameras \cite{Kowdle_2019} have been explored, although they cannot meet the computational requirements of real-time applications. On the other hand, event-cameras achieve high recording speed without saturating bandwidth and resources. 
For these reasons, human pose estimation performed with event-cameras is both interesting and challenging for the community. Approaches that extrapolate information from single events would be ideal, as these methods allow to exploit the interesting advantages of event-cameras. Initial proposals leveraged event-cameras to match events with known objects in the scene \cite{valeiras16_neurom_event_based_pose_estim, Kim_2016}. Rebecq \etal exploit a similar caveat \cite{Rebecq2018EMVS:Real-Time} to predict semi-dense 3D structure of a scene. More recently, Xu \etal \cite{xu20_event} employ events to (1) track features across frames and (2) enhance the intensity outputs of a DAVIS camera. Next, they predict human-poses with VNect \cite{Mehta2017VNect} and Openpose \cite{openpose} models and optimize a multi-step optimization scheme to refine the prediction. Despite its efficiency, their approach relies on a heavy pre-processing phase to extract 3D mesh of the subjects and involves multiple components, each with its own hyper-parameters.
Instead of processing events in small batches, numerous works accumulate events into tensors representations, conducing events in the realm of synchronous deep learning models \cite{Gallego_2020}. To predict human poses through event-cameras, previous works aggregate events to predict 2D poses from multiple point of views and finally triangulate subjects' 3D poses \cite{Calabrese_2019_CVPR_Workshops,baldwin2021timeordered}.
On the other hand, our approach is the first attempt to estimate 3D human pose based on a single DVS camera. We prove that human pose estimation from event-only DVS camera is feasible. For an in-depth discussion of event-cameras and their applications, we refer to the excellent event-cameras summary \cite{Gallego_2020}.

\textbf{Datasets for event-based Human Pose Estimation.}
Few datasets have been recorded using event-cameras, especially if compared with the huge amount of RGB datasets. For human pose estimation, Calabrese \etal released DHP19, a dataset with recordings of 17 subjects and 33 movements. On the other hand, simulating events from RGB videos is a promising path of research, especially since multiple works proved the soundness of training on simulated events. Mueggleret al. \cite{mueggler17_event_camer_datas_simul} generate synthetic events from RGB images and compare real and synthetic events for ego-pose estimation in various scenarios. More recently, simulated events have been employed for image reconstruction \cite{rebecq20_high_speed_high_dynam_range}, depth estimation \cite{gehrig2021combining}, and motion segmentation \cite{StoffregenEvent-BasedCompensation}, especially in high-speed scenarios where RGB ground-truth are hard to collect. In this work, we propose a pipeline to generate synthetic events from the Human3.6m dataset \cite{ionescu11_laten, ionescu14_human} and compare our approach with standard RGB methods to establish a strong benchmark for further research.

\section{Method}
Our goal in this paper is to fill the gap between RGB-based and event-based monocular human pose estimation. In particular, we propose an end-to-end pipeline to predict the skeleton of a subject from the stream of a single event-camera. Figure \ref{fig:abstract} provides an overview of our methodology. An event-camera collects an asynchronous stream of events of a subject moving in the scene. Instead of tracking events as previous works \cite{xu20_event}, we aggregate them into tensor-like frames. Next, we predict three heatmaps planes of the cuboid surrounding the subject and finally build his final 3D pose through triangulation. 

\textbf{Events.} 
Event-cameras have peculiar pixel sensors that capture information
asynchronously. In particular, event-cameras have no central clock; each pixel
senses the light variations of the scene independently according to
\begin{equation}
\label{eq:events}
\begin{split}
  \Delta L(x_k, t_k) > p_k C, \text{ where } \\
  \Delta L(x_k,t_k) \doteq L(x_k,t_k) - L(x_k, t_k - \Delta t_k),
  \end{split}
\end{equation}
where at each pixel $x_k$ we compute the difference in light intensity $\Delta L(x_k,t_k)$ between the current and previous time instance every $\Delta t_k$ seconds. If this difference exceeds a fixed threshold $C$, the pixel emits an event.
An event-camera stream is a sequence of events, each characterized by the
image coordinate pair $(x, y)$, a polarity (related to a positive or a negative change
of intensity), and a timestamp. 

\textbf{Events aggregation.}
Instead of relying on raw asynchronous events, recent literature has shifted toward aggregating events together to build synchronous events representation. 
Common approaches range from simply integrating batch of events (constant-count) to representations involving stochastic modelling of events \cite{Sironi2018HATS:Classification} and temporal sparsity \cite{10.1007/978-3-030-11024-6_54}. As temporal information is critical in 3D human pose estimation \cite{chen20_monoc_human_pose_estim}, our first question is to understand if 3D Human Pose Estimation benefits from specific spatio-temporal representations.
To provide an answer, we compare constant-count representation with spatio-temporal voxel grids \cite{10.1007/978-3-030-11024-6_54}.
While constant-count simply aggregates a constant number of events into an image, spatio-temporal voxel-grid preserves the timestamp contribution of events by building $B$ temporal bins and have been already adopted in image reconstruction \cite{rebecq20_high_speed_high_dynam_range, Scheerlinck_2020_WACV} and depth estimation \cite{gehrig2021combining}. Given a set of N events $\{(\mathbf{x_k}, t_k, p_k)\}_{k=0 \dots N}$, we compute $t_k^*$ as the normalized timestamp of event $k$ into range $[0, B-1]$. Each event $(\mathbf{x_k}, t_k, p_k)$ contribute to each bin $B$ of voxel $V$ proportionally to its normalized timestamp $t_k^*$, as: 

\begin{equation}
        \label{eq:voxel}
            \begin{split}
        \text{V}(\mathbf{x}, t) = \sum_{k=0}^N p_k \max(0, 1 - |t - t_k^*|),
    \\
    \text { where } t_k^* \doteq \frac{B-1}{t_N - t_0}.
          \end{split}
\end{equation}
We set $N=7500$ for both representations and $B=4$ for spatio-temporal voxel-grid.
 
\textbf{Skeleton normalization and projection.}
\begin{figure}[]
    \begin{subfigure}[t]{0.30\textwidth}
    \centering
    \includegraphics[width=\textwidth]{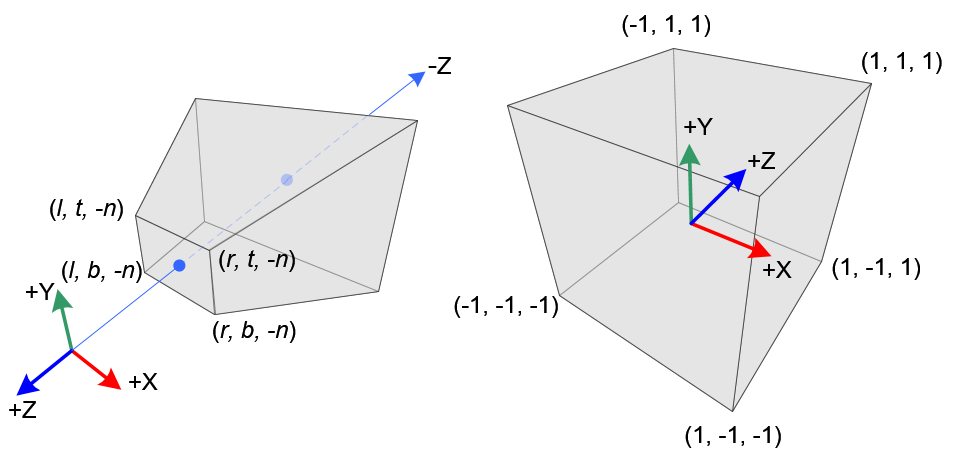}
    \caption{\label{fig:gl}}
    \end{subfigure}
    \hspace{0.01\textwidth}
    \begin{subfigure}[t]{0.15\textwidth}
    \centering
    \includegraphics[width=\textwidth]{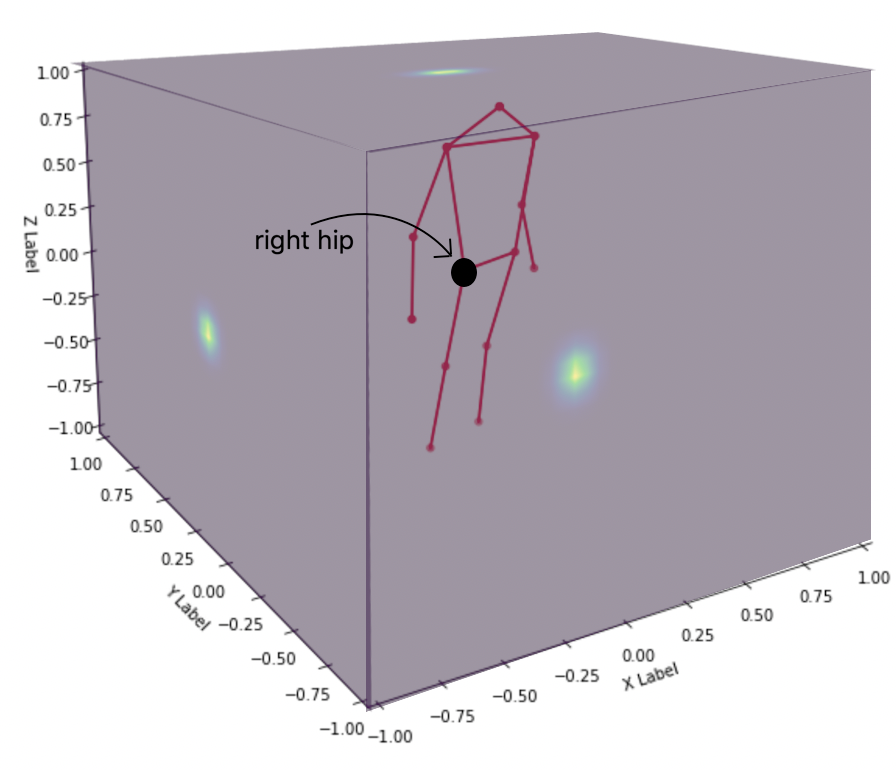}
    \caption{\label{fig:hms}}
    \end{subfigure}
    \caption[Skeleton projection into NDC coordinates]{\label{fig:norm_hm}(a) We define the canonical 3D skeleton pose into a normalized cube \cite{MCREYNOLDS200519, opengl} and reproject the cube into the camera image plane using camera calibration parameters. (b) For each joint, our method extracts the three orthogonal faces of the cube to generate three \textit{marginal} heatmaps.}
\end{figure}
Instead of regressing 3D joints directly, our method relies, as a proxy, on their 2D projections onto specific planes \cite{Nibali20183DHeatmaps}. We generate ground-truth as follows.
First, we project the coordinates $p_{xyz}$ of a joint on a plane parallel to the image plane and placed at depth $z_\text{ref}$ (we adopt the $z$ value of the \textit{head joint} as $z_\text{ref}$). After that, we map the space to a normalized cube $p_{xyz}^\text{NDC}$ (Normalized Device Coordinate - NDC \cite{MCREYNOLDS200519, opengl}): the three coordinates assume values in the range [-1, 1], as in Figure \ref{fig:gl}. Last, we project $p_{xyz}^\text{NDC}$ onto the three orthogonal faces of the cube and blur the projection on each face with a Gaussian Filter to generate ground-truth \textit{marginal heatmaps} $H_{xy}, H_{zy}$ and $H_{xz}$ (Figure \ref{fig:hms}).

\begin{figure}[b]
    \centering
    \begin{subfigure}[]{\linewidth}
    \centering
    \includegraphics[width=.8\linewidth]{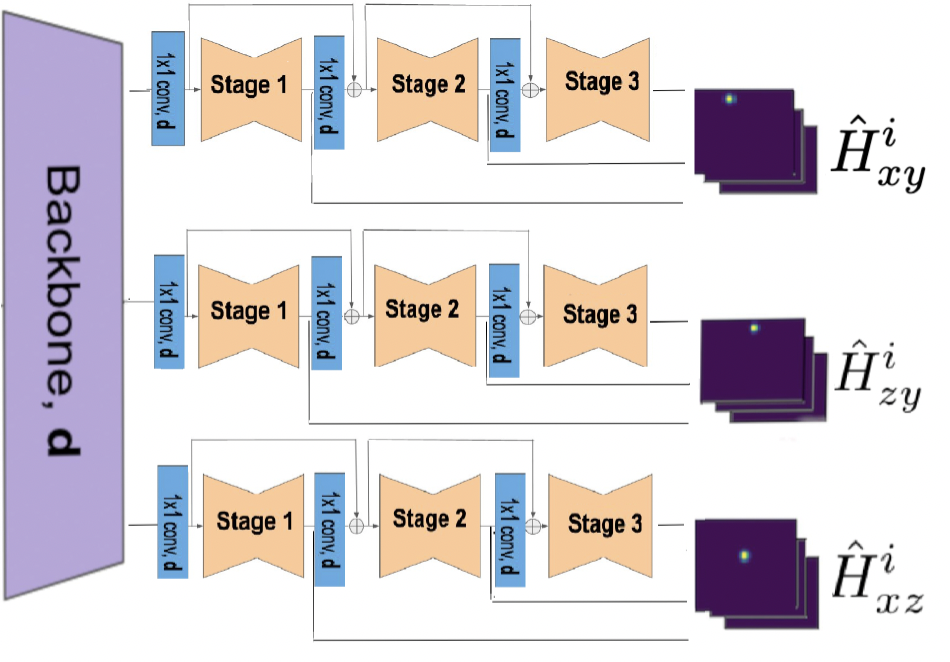}
    \caption{Overview of our model}
    \end{subfigure}
    \hfill
    \begin{subfigure}[]{\linewidth}
    \centering
    \includegraphics[width=.8\linewidth]{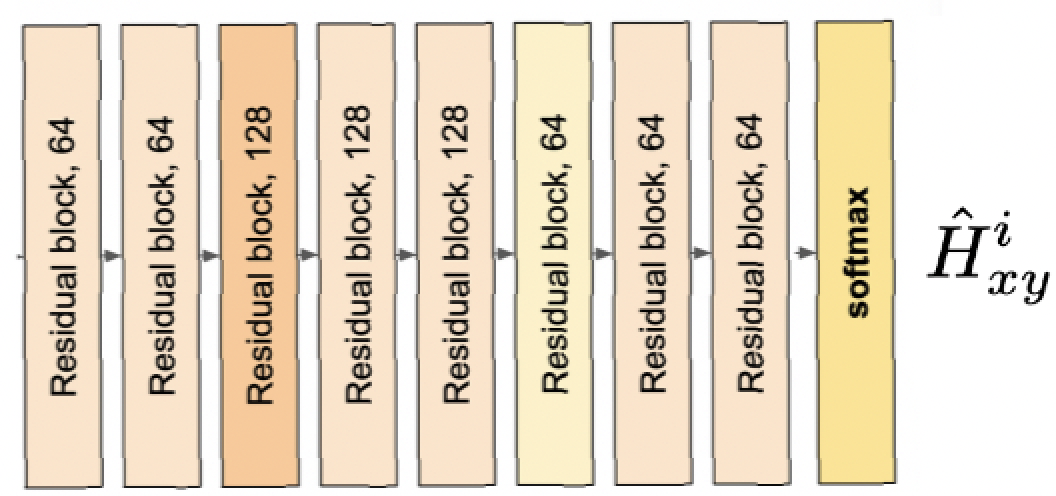}
    \caption{Overview of one stage.}
    \end{subfigure}
    \caption[Model overview]{\label{fig:approach}(a) We process event-frames with a \textbf{backbone} that outputs features of depth \textbf{d}. Next, we adopt three sequential stages to output $3 \times J$ intermediate heatmaps. We apply an intermediate loss to each stage \cite{Neil2016PhasedSequences} and accumulate the losses to solve the vanish gradient problem. (b) Each stage process its input with three deep Convolutional Neural Network through an auto-encoder architecture.}
\end{figure}

\textbf{Predicting marginal heatmaps.}
We design our approach upon marginal heatmaps \cite{mehta17_monoc_human_pose_estim_wild, Nibali20183DHeatmaps} and first predict three 2D heatmaps from our monocular input. Figure \ref{fig:approach} summarizes our model. We first process the event-frame input with a backbone to extract intermediate representations. In particular, we adopt a ResNet-34 \cite{DBLP:journals/corr/HeZRS15} which is cut after the second residual block. The feature extractor initialization is a critical design choice of our approach and we experimentally ablate possible alternatives in Section \ref{subsec:ablation}, where we compare different initialization and pre-training strategies and provide evidence of the benefits of RGB-to-events transfer learning. 
The main model consists in three branches, one for each marginal projection ($xy$, $zy$, and $xz$). Each branch is further made of three stages (Figure \ref{fig:approach}(a)), each consisting in a hourglass-like CNN, as detailed in Figure \ref{fig:approach}(b). For each stage we compute an intermediate loss. The result of each stage is also aggregated (summation) with the previous output to feed the next stage, in a residual-like fashion. According to \cite{Newell2016StackedEstimation}, intermediate losses help alleviating the problem of vanishing gradients.

\textbf{Aggregating marginal heatmaps.} 
Our model is trained jointly to predict the intermediate heatmaps as well as the normalized skeletal coordinates. We apply the soft-argmax operator \cite{Neil2016PhasedSequences} to extract the normalized coordinates of each joints onto the $xy$, $xz$, and $yz$ planes. We choose the predictions from the $xy$-plane for the $xy$ coordinate of the final prediction $\hat p_{xyz}$, as they match naturally with the input image. For $z$, we average the $zy$ and $xz$ predictions. Eq. \ref{eq:pred} summarizes these steps as: 
\begin{equation}
            \begin{split}
    \hat{H}^i_{xy}, H^i_{xz}, \hat{H}^i_{yz} = \text{Model}(\mathbf{x}) \\
    \begin{bmatrix}x^i_{xy}, y^i_{xy}\end{bmatrix}= \text{soft-argmax}(\hat{H}^i_{xy}) \\
    \begin{bmatrix}x^i_{xz}, z^i_{xz}\end{bmatrix}= \text{soft-argmax}(\hat{H}^i_{xz}) \\ 
    \begin{bmatrix}y^i_{zy}, z^i_{zy}\end{bmatrix}= \text{soft-argmax}(\hat{H}^i_{zy}) \\
        \hat p^i_{xyz} = \begin{bmatrix}x^i_{xy}, y^i_{xy}, \frac{z^i_{xz} + z^i_{zy}}{2}\end{bmatrix}.
                \end{split}
                   \label{eq:pred}
\end{equation}

\textbf{Losses.} As the full pipeline is differentiable, we can back-propagate the geometrical error between joints predictions and ground-truths and train our model end-to-end. Moreover, we can interpret marginal heatmaps as probability distributions of joints locations. In this framework, we apply the Jensen–Shannon divergence (Equation \ref{eq:jsd_loss}) between predicted heatmaps $\hat H^i$ for stage $i$ and ground-truth heatmaps $H$. $\text{JSD}$ is based on the Kullbeck-Leibler divergence ($\text{KL}$), it is symmetric and has only finite values given by:
\begin{equation}
    \text{JSD}(H, \hat H) = \frac{1}{2} \text{KL}(H \| \hat H) + \frac{1}{2} \text{KL}(\hat H \| H).
          \label{eq:jsd_loss}
\end{equation}   
The Jensen-Shannon divergence and the geometrical loss for each stage $i$ are aggregated into the final loss $L$ as:
\begin{equation}
    \begin{split}
        L=\sum_i L_\text{geometrical}(\hat p_\text{xyz}^i, p_\text{xyz}) + \text{JSD}(H_{xy}, \hat H_{xy}^i) + \\ \text{JSD}(H_{xz}, \hat H_{xz}^i) + \text{JSD}(H_{zy}, \hat H_{zy}^i), \\
                \end{split}
    \label{eq:final_loss}
\end{equation}
where $L_\text{geometrical}(\hat p_\text{xyz}^i, p_\text{xyz}) = \lVert \hat p_\text{xyz}^i - p_\text{xyz}\lVert_2$.

\section{Experiments}
We test our approach on our novel Event-Human3.6m dataset and provide extensive comparison to support our claims. Moreover, we experiment on real events from the event-based DHP19 dataset. For both the dataset, we address the scale-depth ambiguity using a ground-truth depth point and calculate the Mean Per-Joint Precision Error (MPJPE) between the de-normalized predictions and the ground-truths \cite{Nibali20183DHeatmaps, Luvizon_2D3D_Pose_Estimation_CVPR_2018_paper}.

\subsection{Datasets}
\textbf{DHP19 dataset.}
DHP19 \cite{Calabrese_2019_CVPR_Workshops} contains 33 recordings of 17 subjects of different sex, age, and size. Each subject is recorded with four DVS cameras from different angles. Nevertheless, the range of movements in the recordings is narrow. Most of the activities, such as \textit{legs kicking} and \textit{arms abductions}, are conducted on the spot, with the exception of \textit{slow jogging} and \textit{walking}. Moreover, few recordings spot real life activities. These gaps in the data limit its applications in real scenarios.

\begin{figure}[ht] \centering

  \begin{subfigure}[b]{.32\linewidth} \centering
    \includegraphics[width=\linewidth]{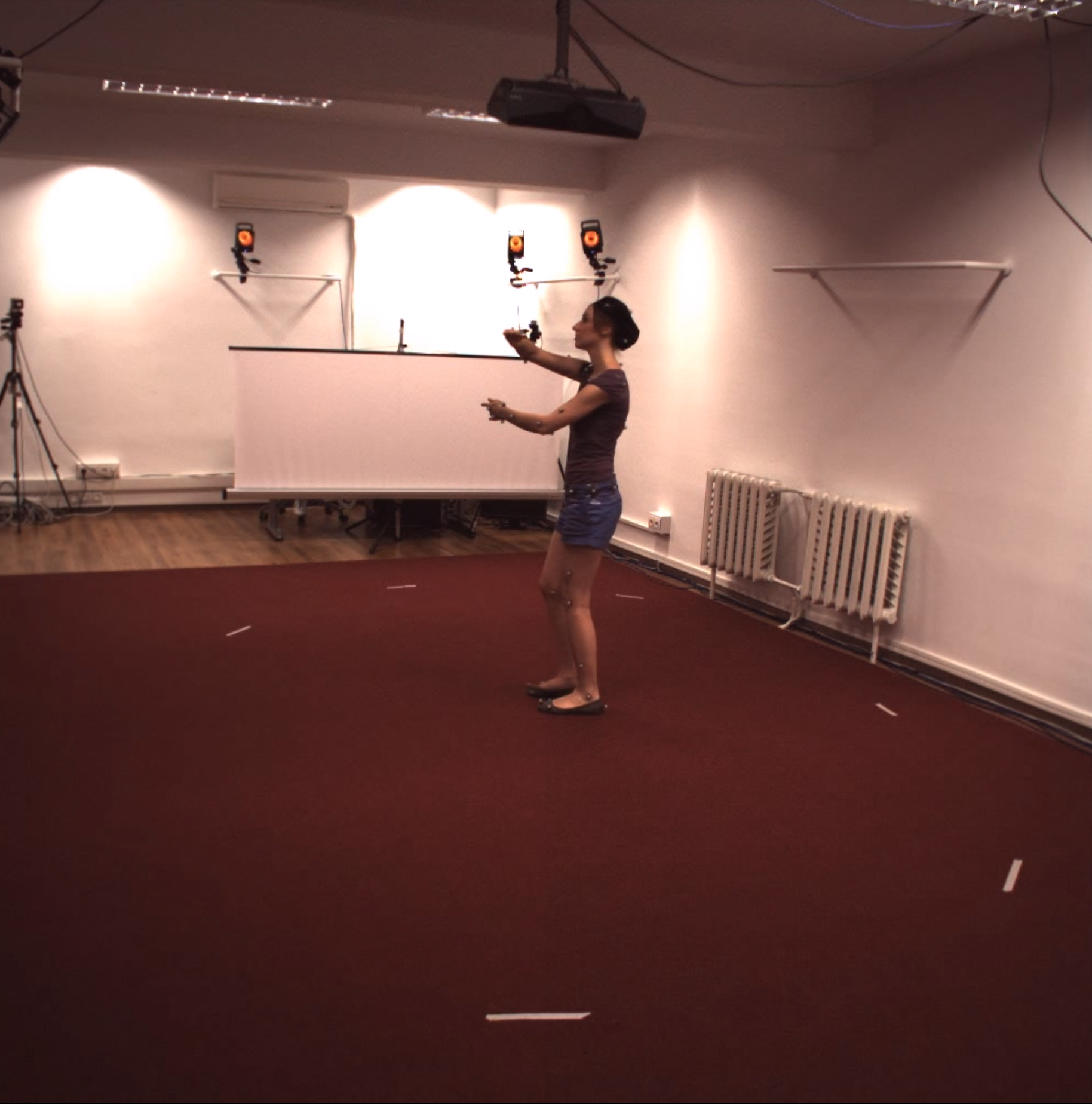}
      \caption{}
  \end{subfigure} 
    \begin{subfigure}[b]{.25\linewidth} \centering
    \includegraphics[width=\textwidth]{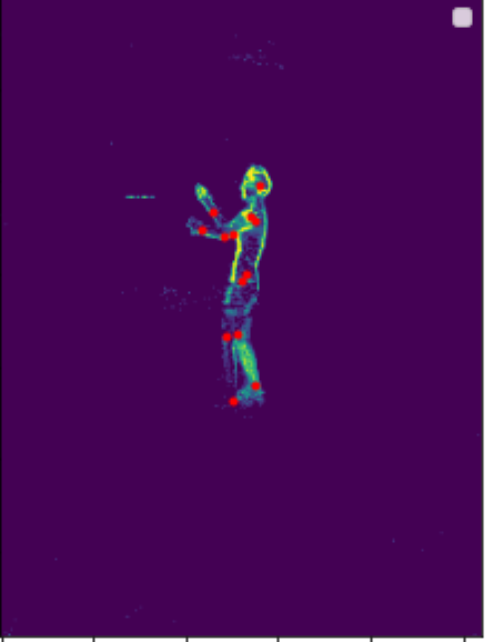}
      \caption{}
    
  \end{subfigure} 
    \begin{subfigure}[b]{.38\linewidth} \centering
    \includegraphics[width=\textwidth]{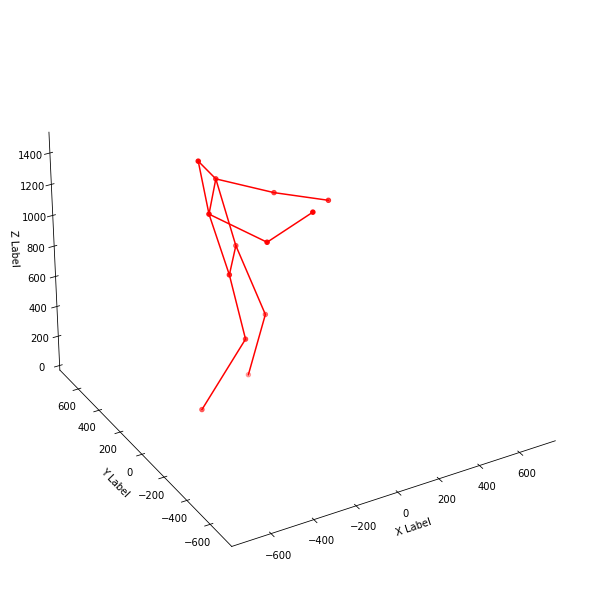}
      \caption{}
  \end{subfigure} 
  \hfill
 
\caption[]{\label{fig:h3m}We simulate raw events from Human3.6m recordings (a) with the open-source simulator ESIM-Py \cite{Rebecq2018ESIM:Simulator}. We set the simulators parameters $cp=cn=0.2$, $\text{log-eps} = 1e{-3}$, and $\text{refractory-period}=1e{-4}$, as this setting produces synthetic events similar to DHP19 event-cameras recordings. Next, we accumulate events into event-frames (b) and interpolate ground-truths to match  timestamps (c).}
\end{figure}

\textbf{Event-Human3.6m dataset.}
In the previous section we highlight some limitations of the DHP19 dataset \cite{Calabrese_2019_CVPR_Workshops}, especially related to the narrowness of movements and activities that it provides. To solve these gaps, we contribute with a new simulated datasets based on the Human3.6m dataset \cite{ionescu11_laten,ionescu14_human}. Human3.6 recordings include 11 subjects and different activities from real scenarios, such as \textit{walking with a dog}, \textit{talking at the phone}, and \textit{giving directions}. Consequently, extensive research has adopted the standard Human3.6m dataset to evaluate monocular Human Pose Estimation methods \cite{Nibali20183DHeatmaps, Mehta2017VNect, pavlakos17_coars_fine_volum_predic_singl}. We believe event-based research will benefit from our Event-Human3.6m, as it extends DHP19 with more challenging scenarios and provides a new benchmark for monocular human pose estimation algorithms.
We adopt the ESIM-Py simulator \cite{Rebecq2018ESIM:Simulator} to convert the RGB recordings of Human3.6m into events and synchronize raw joints ground-truth with events frames through interpolation (Figure \ref{fig:h3m}). As a result, Event-Human3.6m and DHP19 have comparable ground-truths and event frames. In the following sections we reports extensive experiments on both DHP19 and Event-Human3.6m to test the benefits of our proposal.
 
\textbf{Training details.}
We explore different hyper-parameters settings empirically. In the following experiments, we train our method on 4 Tesla V100 16Gb GPUs and adopt a batch size (per GPU) of $32$. For updating the gradients, we opt for Adam optimizer \cite{kingma2017adam} with learning rate of $0.0003$. We interrupt the train at local convergence through an early-stopping strategy. We evaluate our approach with 1 stage (7M of parameters, 91 MB of storage) and 3 stages (21M parameters, 300MB of storage).

\subsection{Results}
Here we discuss the performance of our approach and validate it on the two datasets.

\textbf{Evaluation on DHP19.}
\label{exp:dhp19}
We test 1-stage and 3-stage models with spatio-temporal voxel-grid and constant-count representations. Table \ref{tab:dhp19} reports the Mean Per Joint Precision Error (MPJPE, in mm) and summarizes the results. As reference, we compare to the stereo approach of Calabrese \etal \cite{Calabrese_2019_CVPR_Workshops}. As a first observation, our methodology performs only slightly worse than the stereo approach (13mm difference). In this setting, constant-count representation performs better than voxel-grid. In the ablations, we elaborate on the differences between the two representations when different backbones are adopted as feature extractors. Moreover, we provide results for our single stage and 3-stages model and compare them. Table \ref{tab:dhp19} shows that multiple stacked stages and intermediate losses provide sensible performance benefits, at the cost of an increase in computational costs and model size. 

\begin{table}[h]
\caption{\label{tab:dhp19}Results refer to DHP19 dataset \cite{Calabrese_2019_CVPR_Workshops}. We compare our approach with 1 and 3 stack of stages across constant-count and voxel-grid representation.}
\centering
\begin{tabular}{lrr}
\hline
\hline
 Method & input & MPJPE(mm)\\
 \hline
Calabrese \etal \cite{Calabrese_2019_CVPR_Workshops} & stereo & 79.63  \\
Constant-count -- stage 3 & monocular & 92.09\\
Voxel-grid -- stage 3 & monocular & 95.51 \\
Constant-count -- stage 1 & monocular & 96.69\\
Voxel-grid -- stage 1 & monocular & 105.24\\
\hline
\end{tabular}
\end{table}

\textbf{Evaluation on Event-Human3.6m.}
For each subject, we keep 13 out the 32 provided joints to build skeletons that are compatible with DHP19 ground-truths and evaluate our approach on a cross-subject protocol. We train our models on subjects 1, 3, 5, 7, 8 and test on subjects 9 and 11. Similar works \cite{Nibali20183DHeatmaps, Luvizon_2D3D_Pose_Estimation_CVPR_2018_paper, pavlakos17_coars_fine_volum_predic_singl} evaluate monocular approaches on every $64^{th}$ frame of the recordings. We adapt this evaluation protocol to our asynchronous Event-Human3.6m by taking event-frames corresponding to the same testing frames. Table \ref{tab:h3m} reports the results of our approach with constant-count and voxel-grid representations. Moreover, we compare our methodology to state-of-the-art RGB approaches \cite{Kanazawa_2018, Nibali20183DHeatmaps, pavlakos17_coars_fine_volum_predic_singl, Luvizon_2D3D_Pose_Estimation_CVPR_2018_paper}. Despite the gap with standard computer-vision techniques, our approach performs fairly against existing RGB approaches.

\begin{table}[h]
\caption{\label{tab:h3m}Comparison between RGB approaches on Human3.6m and our approach on its synthetic counterpart. We adopt a standard cross-subject protocol to validate on the same testing strategy as RGB approaches.}
\centering
\resizebox{\columnwidth}{!}{
\begin{tabular}{llr}
\hline
\hline
 Method & input & MPJPE(mm)\\
 \hline
 Metha \etal \cite{Mehta2017VNect} (ResNet-50) & RGB & 80.50\\
 Kanazawa \etal  \cite{Kanazawa_2018} & RGB & 88.00\\
 Nibali \etal \cite{Nibali20183DHeatmaps} & RGB & 57.00\\
 Pavlakos \etal \cite{pavlakos17_coars_fine_volum_predic_singl} & RGB & 71.90\\
 Luvizon \etal \cite{Luvizon_2D3D_Pose_Estimation_CVPR_2018_paper} & RGB & 53.20\\
 Cheng \etal \cite{Cheng_2020} & RGB & \textbf{40.10}\\
\hline
\hline
Spatio-temporal voxel-grid \textbf{(Ours)} & Events & 119.18\\
Constant-count \textbf{(Ours)} & Events & 116.40\\
\hline
\end{tabular}}
\end{table}

\subsection{Ablation study}
\label{subsec:ablation}
In this Section, we deepen different aspects of our approach in more detail. In particular, we are interested to explore \textit{what movements cause our approach to fail} and \textit{how backbone initialization impacts performance}. In the following, we discuss these questions in more details.


\textbf{Transfer learning and pre-training tasks.} Event representations and RGB images share some commonalities, especially edges and corners. However, if we compare them closely, we find subtle differences, since event-cameras recordings are highly correlated to the dynamic of the scene. If the RGB/event-frames analogy held, event-based vision could benefit widely from advancements in standard computer vision. As an example, recent computer vision research provides strong evidence in support of transfer learning from large dataset, e.g., the ImageNet dataset \cite{imagenet_cvpr09, DBLP:journals/corr/HuhAE16}. Further works explore and validate the correlation between 3D Human Pose Estimation and reconstruction tasks \cite{taskonomy2018}. These insights are supported by common intuition, as both tasks involve an understanding of the structure of the scene. Despite the differences between event and standard cameras, recent works validate the transfer learning hypothesis from RGB to constant-count representation \cite{Maqueda_2018} and learnable representations \cite{gehrig2019end}. Moreover, Rebecq \etal provide evidence for direct transfer learning by predicting natural images from spatio-temporal event-frames \cite{rebecq20_high_speed_high_dynam_range}. 

Our work contributes further to this line of research with two evaluations. First, we compare ImageNet and random initialized models for solving monocular human pose estimation with both constant-count and voxel-grid representations. Second, we attempt to validate if different pre-training tasks help with event-based Human Pose Estimation. For this purpose, we train an auto-encoder consisting of a ResNet-34 as encoder and a small DeconvCNN as decoder. For comparison, we train a ResNet-34 and a ResNet-50 CNN on action recognition task, which has lower correlation with human pose estimation.
Next, we test our approach with 4 backbones (random-initialized, action recognition task, reconstruction task, and ImageNet initialized) and compare the results on DHP19 dataset. Table \ref{tab:abl_backbone} reports the MPJPE for both constant-count and voxel-grid representations. Constant-count frames benefit more from standard computer vision, especially from ImageNet transfer-learning. In fact, our model with ImageNet-pretrained ResNet34 outperforms all others approaches when we adopt constant-count representation. 

Spatio-temporal frames have few similarities with standard RGB images; in fact, it is unclear if this approach can benefit from ImageNet transfer learning. Our experiments reflects these differences, as ImageNet pretrained ResNet-34 and ResNet-50 backbones have lower performance than the random-initialised counterpart.

We discuss Table \ref{tab:abl_backbone} to explore further if recent research in pre-training tasks \cite{taskonomy2018} is valid in event-based vision. Despite the correlations evidences in RGB settings, we find that auto-encoders backbones are performing worse than the classification counterpart; this conclusion is valid from both representations. Indeed, action-recognition pre-training emerges favorably, especially for spatio-temporal voxel-grid. Our interpretation is that pre-training assumptions fail because of the spatial sparsity of event-representations. Further research is mandatory to unlock better pre-training strategies for event-based vision.

\begin{table}[ht]
\caption{\label{tab:abl_backbone} We report the Mean Per Joint Precision Error (MPJPE, in mm) of our 3-stages approach equipped with different initialization strategies. ResNet-34 with ImageNet initialization emerges favorably for constant-count representation. Moreover, we find no benefits in adopting a reconstruction task as pre-training task, although standard computer vision research suggests the opposite\cite{taskonomy2018}.}
\centering
\small
\begin{tabular}{lllr}
\hline
\hline
 Repr. & Model & Initialization & MPJPE (mm) \\
 \hline
\multirow{8}{*}{\rotatebox{90}{constant-count}} &\multirow{4}{*}{ResNet-34} 
& Random initialized & 92.22\\
&& Action recognition & 95.19\\
&& Reconstruction& 98.89\\
&&ImageNet & \textbf{92.09}\\\cline{3-4}
&\multirow{3}{*}{ResNet-50} 
&Random initialized & 92.22\\
&&Action recognition & 92.26\\
&& ImageNet & 92.51\\
\hline
\hline
\multirow{8}{*}{\rotatebox{90}{voxel-grid}} &\multirow{4}{*}{ResNet-34} 
&Random initialized & 93.06\\
&&Action recognition & 95.26\\
&&Reconstruction & 105.44\\
&&ImageNet & 95.51\\\cline{3-4}
&\multirow{3}{*}{ResNet-50} 
&Random initialized & 93.88\\
&&Action recognition & 93.54\\
&& ImageNet & 93.98\\
\hline
\end{tabular}
\end{table}

\textbf{Per-movements comparison.} Events are highly coupled with the dynamic of the scene. If parts of the body are static, fewer events are recorded. As a consequence, spatial sparsity increases and makes prediction tasks more challenging. To evaluate the impact of static body parts on our approach, we propose a per-movements study for our ImageNet-pretrained method. Table \ref{tab:per_movements} compares our constant-count and spatio-temporal voxel-grid approaches with DHP19 \cite{Calabrese_2019_CVPR_Workshops} event-based stereo approach. Differently from \cite{Calabrese_2019_CVPR_Workshops}, our approach is based upon the more recent state of the art solutions \cite{Newell2016StackedEstimation,Nibali20183DHeatmaps} and reaches a higher per-movement accuracy and lower per-movement standard deviation. As expected, performance decreases when subjects perform movements with only parts of the body (e.g., \textit{Punch up forwards left} implies static legs). This drop in performance matches the results of the stereo-vision approach (e.g., \textit{Punch forwards left/right}). On the other hand, we notice above average performances for movements that involve the whole body, such as \textit{knee lift} and \textit{hand movements} (during these movements, subjects move on the spot and the whole body generates events). 
 \begin{table}[h]
 \caption{We compare the per-movement MPJPE between ours and DHP19 \cite{Calabrese_2019_CVPR_Workshops} stereo approach. Both fail when parts of the body are static and shine when the scene is more dynamic. In bold we highlight \textbf{worst results per column} while with an underline we show \underline{best results per column.}}
 \centering
 \label{tab:per_movements}
 \resizebox{.99\columnwidth}{!}{
  \begin{tabular}{lrrr}
 \hline \hline 
 & Stereo \cite{Calabrese_2019_CVPR_Workshops} & Voxel-grid & Constant-count\\ \hline \hline 
Left arm abduction & 115.04 & 82.32 & 80.41\\
Right arm abduction & 99.65 & 81.92 & 79.68\\
Left leg abduction & 84.65& 110.07 & 105.39\\
Right leg abduction & 78.35& 99.87& 93.81\\
Left arm bicep curl &103.29 & 90.49 & 86.40\\
Right arm bicep curl & 121.06& 80.75 & 95.73\\
Left leg knee lift &74.97 &\underline{71.60} & \underline{72.14}\\
Right leg knee lift & 71.95& 78.47 & \underline{72.49}\\
Walking 3.5 km/h &58.75 & 86.88 & 84.74\\
Single jump up-down & 82.23& 80.11 & 76.73\\ 
Single jump forwards & 80.53&  89.92 & 85.10\\
Multiple jumps & \underline{53.57}&  99.47 & 93.83\\
Hop right foot & \underline{55.56}& 89.51 & 84.16\\
Hop left foot & \underline{54.21}&97.86& 91.60\\
Punch forward left & \textbf{148.57}& 114.97 & \textbf{117.87}\\
Punch forward right &\textbf{135.92}& 98.35 & 93.69\\
Punch up forwards left& 111.35& \textbf{124.89}& \textbf{124.81}\\
Punch up forwards right & \textbf{131.46}& 103.01 & 106.56\\
Punch down forwards left &106.92 & 105.98& 105.04\\
Punch down forwards right &98.28 & 90.02 & 89.90\\
Slow jogging & \underline{55.16}& 98.05 & 89.11\\
Star jumps & 76.23& 108.89& 106.77\\
Kick forwards left &  111.66& \textbf{117.92} & 93.07\\
Kick forwards right  & 112.49 & \textbf{117.91} & 109.85\\
Side kick forwards left  & 118.00& \textbf{128.38 }& \textbf{120.39}\\
Side kick forwards right  & 104.67& 115.76 & \textbf{111.86}\\
Hello left hand &  96.22& 89.08& 87.22\\ 
Hello right hand  & 101.32&\underline{71.82}  &\underline{69.83}\\
Circle left hand  & 110.59& 99.17& 95.89\\
Circle right hand  & 112.44& 84.00 & 76.55\\
Figure-8 left hand  & 110.69& 90.95 & 88.10\\
Figure-8 right hand & \textbf{123.59} & \underline{72.42}& \underline{72.49}\\
Clap& 122.93 & \underline{81.03} & 77.77\\
\hline 
Mean (standard deviation) & 98.06 ($\pm 16.60$) & 95.51 ($\pm 15.30$) & 92.09 ($\pm 14.49$)
 \end{tabular}}
 \end{table}
  \begin{figure*}[ht] \centering 
            \begin{subfigure}[b]{.12\linewidth} \centering
    \includegraphics[width=\textwidth]{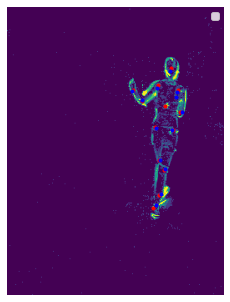}
  \end{subfigure} 
    \begin{subfigure}[b]{.12\linewidth} \centering
    \includegraphics[width=\textwidth]{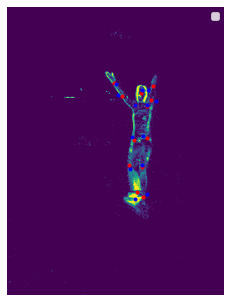}
  \end{subfigure} 
            \begin{subfigure}[b]{.12\linewidth} \centering
    \includegraphics[width=\textwidth]{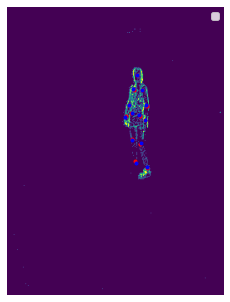}
  \end{subfigure} 
              \begin{subfigure}[b]{.12\linewidth} \centering
    \includegraphics[width=\textwidth,height=2.7cm]{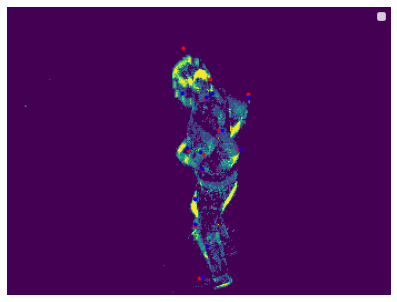}
  \end{subfigure} 
            \begin{subfigure}[b]{.12\linewidth} \centering
    \includegraphics[width=\textwidth]{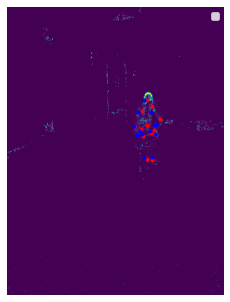}
  \end{subfigure} 
        \begin{subfigure}[b]{.12\linewidth} \centering
    \includegraphics[width=\textwidth]{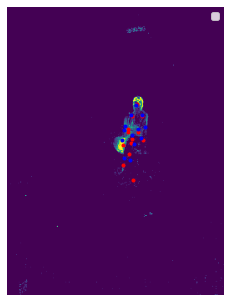}
  \end{subfigure} 
              \begin{subfigure}[b]{.12\linewidth} \centering
    \includegraphics[width=\textwidth,height=2.7cm]{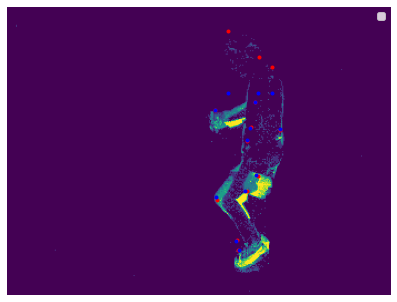}
  \end{subfigure} 
  
  \hfill
  
        \begin{subfigure}[b]{.12\linewidth} \centering
    \includegraphics[width=\textwidth]{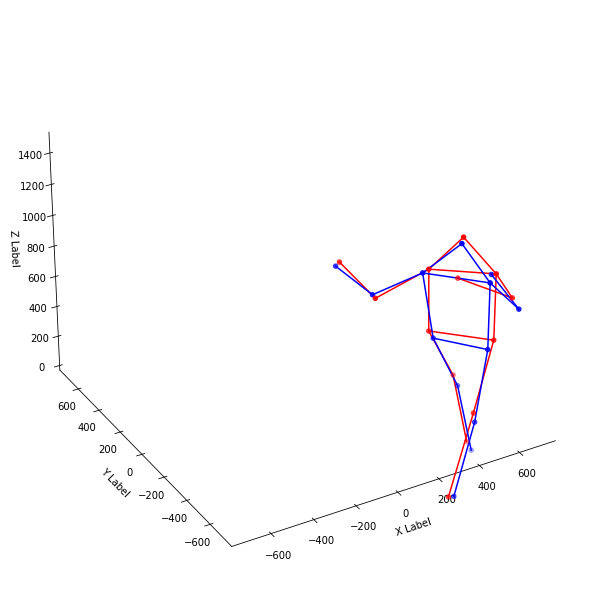}
    \caption{}
  \end{subfigure}
  \begin{subfigure}[b]{.11\linewidth} \centering
    \includegraphics[width=\textwidth]{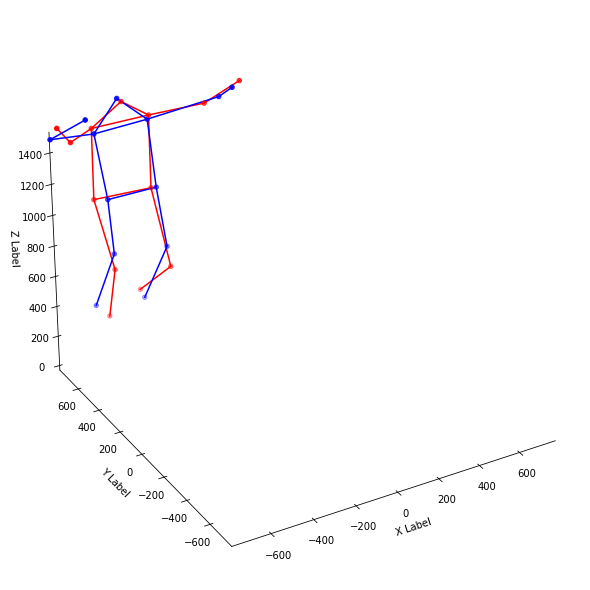}
       \caption{}
  \end{subfigure}
      \begin{subfigure}[b]{.12\linewidth} \centering
    \includegraphics[width=\textwidth]{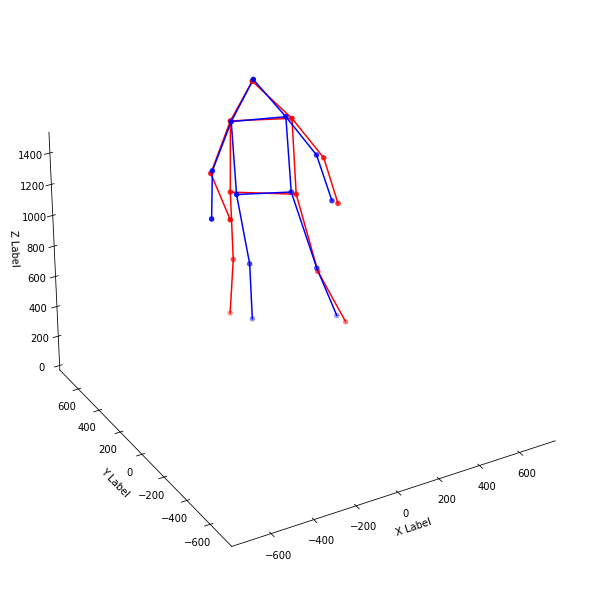}
       \caption{}
  \end{subfigure}
      \begin{subfigure}[b]{.12\linewidth} \centering
    \includegraphics[width=\textwidth]{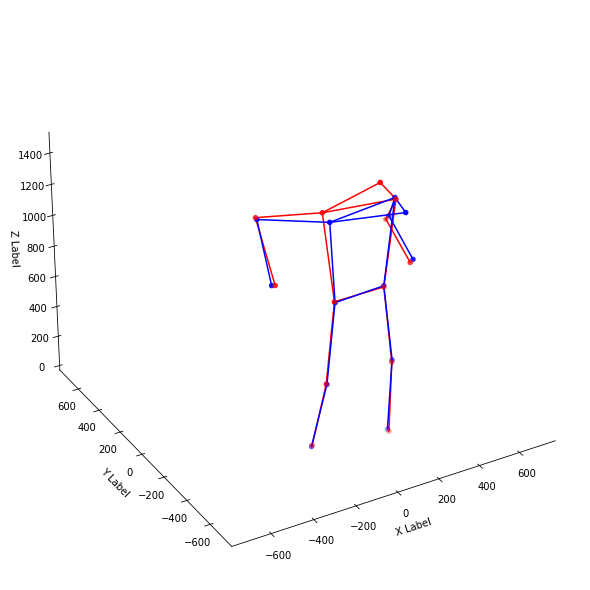}
       \caption{}
  \end{subfigure}
    \begin{subfigure}[b]{.12\linewidth} \centering
    \includegraphics[width=\textwidth]{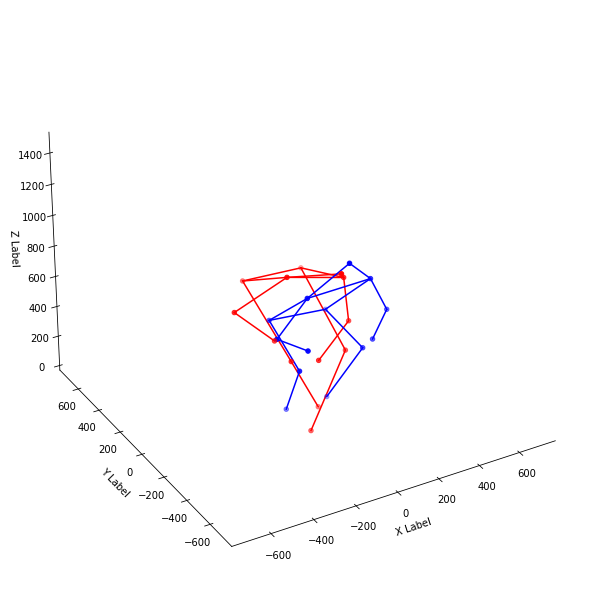}
       \caption{}
  \end{subfigure}
      \begin{subfigure}[b]{.12\linewidth} \centering
    \includegraphics[width=\textwidth]{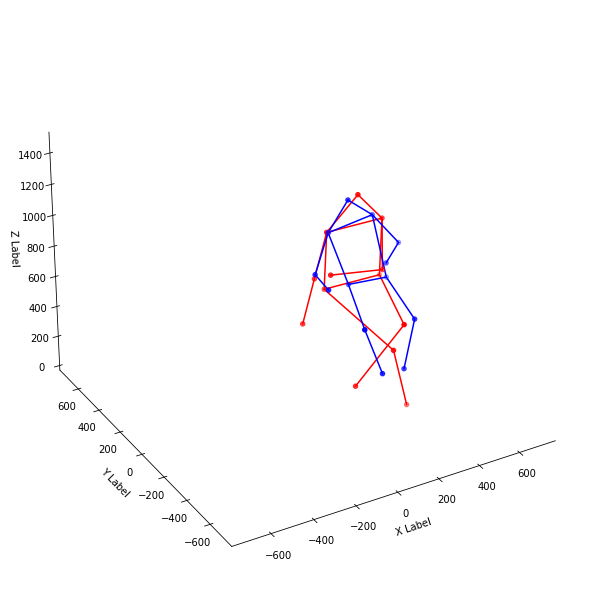}
       \caption{}
  \end{subfigure}
        \begin{subfigure}[b]{.12\linewidth} \centering
    \includegraphics[width=\textwidth]{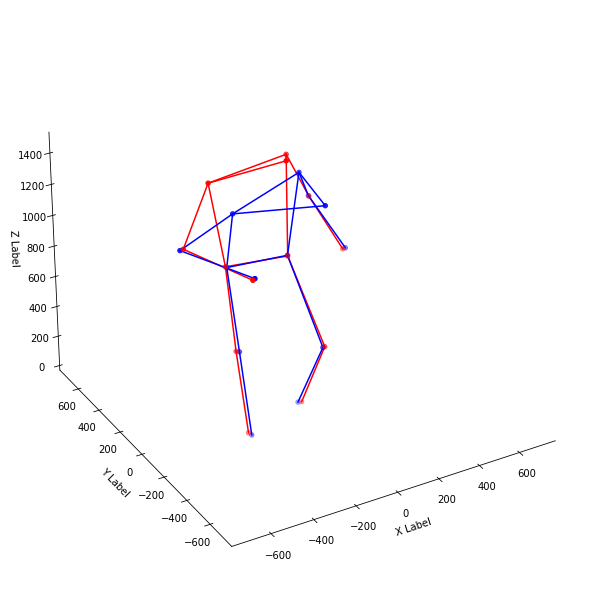}
       \caption{}
  \end{subfigure}
  \caption{\label{fig:qualitative_results}Our approach achieves good performance when subjects are actively moving, as in (a)--(d), but fails to predict the skeletons satisfactorily when some parts of the body remain static during the movements, as in (e)--(g). 
  }
\end{figure*}
\section{Discussion and Conclusions}
We have proposed a deep learning approach for event-based human pose estimation from a single event-camera. Our method aggregates events into synchronous tensor representations to feed a multi-stage Convolutional Neural Network. Our architecture predicts three orthogonal heatmaps which are triangulated to obtain the final 3D pose. 
We validated our approach on the event-based DHP19 dataset, where it showed satisfactory per-movement performance against DHP19 stereo approach \cite{Calabrese_2019_CVPR_Workshops}. Moreover, we proposed Event-Human3.6m, a new dataset of simulated events from the standard Human3.6m \cite{ionescu14_human}. Event-Human3.6m extends DHP19 with more challenging movements and actions. We conducted experiments on the synthetic dataset and adopted a cross-subject protocol which is comparable to the standard RGB testing. Although we recognize the differences between synthetic and RGB datasets, our proposal achieved an accuracy comparable to RGB approaches. These experiments demonstrated the effectiveness of our method.

Figure \ref{fig:qualitative_results} reports challenging examples where our method underperforms. Static parts of the body generated fewer events and are difficult to predict accurately. We leave this issue for further investigations. Next, we conducted extensive ablations studies to understand how event-based vision can benefit from RGB transfer learning and pre-training. Experiments showed that ImageNet pre-training boosts our approach more than pre-training tasks. Moreover, action recognition pre-training task archived higher performances than reconstruction pre-training, although extensive computer vision research suggests the opposite. 
Future research should consider closely the relationships between events and RGB cameras in transfer-learning and multi-task learning settings. Further works to answer these open questions can benefit from our synthetic Event-Human3.6m.

{\small
\bibliographystyle{ieee_fullname}
\bibliography{references}
}

\end{document}